# A Possibility Distribution Based Multi-Criteria Decision Algorithm for Resilient Supplier Selection Problems

**Dizuo Jiang, Md Mahmudul Hassan, Tasnim Ibn Faiz, and Md. Noor-E-Alam**[*]

Department of Mechanical and Industrial Engineering,
Northeastern University
360 Huntington Ave, Boston, MA 02115, USA

*Corresponding author email: mnalam@neu.edu

## Abstract

Thus far, limited research has been performed on *resilient supplier selection*—a problem that requires simultaneous consideration of a set of numerical and linguistic evaluation criteria, which are substantially different from traditional supplier selection problem. Essentially, resilient supplier selection entails key sourcing decision for an organization to gain competitive advantage. In the presence of multiple conflicting evaluation criteria, contradicting decision makers, and imprecise decision relevant information (DRI), this problem becomes even more difficult to solve with the classical optimization approaches. Possibility distribution based Multi-Criteria Decision Analysis (MCDA) is a viable alternative approach for handling inherent uncertainty of imprecise DRI associated with the evaluation offered by a group of contradicting decision makers. However, prior research focusing on MCDA based supplier selection problem has been lacking in the ability to provide a seamless integration of numerical and linguistic evaluation criteria along with the consideration of multiple decision makers. To address these challenges, we present a comprehensive decision-making framework for ranking a set of suppliers from resiliency perspective. The proposed algorithm is capable of leveraging imprecise and aggregated DRI obtained from crisp numerical assessments and reliability adjusted linguistic appraisals from a group of decision makers. We adapt two popular tools - Single Valued Neutrosophic Sets (SVNS) and Interval-valued fuzzy sets (IVFS), and for the first time extend them to incorporate both crisp and linguistic evaluations in a group decision making platform to obtain aggregated SVNS and IVFS decision matrix. This information is then used to rank the resilient suppliers by using TOPSIS method. We present a case study to illustrate the mechanism of the proposed algorithm. A sensitivity analysis demonstrates the strength of the proposed algorithm to generate alternative ranking scheme with respect to the priorities of evaluation criteria, and thus shows the potential to provide a reliable decision-making framework.

# 1. Introduction

Due to the competitive nature of the current open market economy, a manufacturer's ability to avoid and/or absorb disruptions in supply chain has become a crucial requirement to survive and thrive in the market. The flow of material and information through the present-day supply chains can be disrupted by diverse unpredictable natural catastrophes such as earthquakes, floods, hurricanes, or unexpected man-made disasters such as labor strikes, bankruptcy or terrorist attack (Sawik, 2013). These disruptive events with low probability of occurrences can cause huge financial impact on supply chain operations. The Japanese earthquake occurred in 2011 struck the supply chains of motor vehicle companies such as General Motors (GM) and Toyota (Canis, 2011). The unexpected disaster resulted in substantially reduced production capacity in the U.S., while full restoration of the supply chain capacity took months. As a small example, the earthquake disrupted the operations of Renesas Electronics, manufacturer of 40% of the world's supply of automotive microcontrollers; the negative impact of such disruption was experienced by the worldwide automotive industry. To avoid or absorb the disruptions caused by disasters, either natural or anthropogenic, a manufacturer must design its supply chain network to be resilient. To a great extent, resilience capacity of a supply chain is preserved by resilient suppliers. Yossi (Sheffi, 2005) and Rice (Sheffi & Rice Jr, 2005) have introduced the definition of supply chain resilience and resilient supplier characteristics. A resilient supplier has high capability to resist or absorb disaster impact and can get back to usual performance quickly following a disaster. To select an optimal resilient supplier among multiple alternatives, several aspects need to be taken into consideration.

Supplier selection is a complicated, yet crucial decision problem because of its far-reaching influence on the quality, efficiency and reliability of the supply chain. The decision problem involves weighing several alternatives against multiple conflicting criteria. This becomes even more complicated when evaluation is done by multiple decision makers, each using their own perceptions about importance of criteria and the performance of the alternatives. Furthermore, pertinent information is often imprecise and available in linguistic form. *Multi Criteria Decision Analysis* (MCDA) is one of the most promising approaches to solve this complex decision-making process, as addressed by several researchers (Devlin & Sussex, 2011; Gregory et al., 2012; Hasan, Shohag, Azeem, & Paul, 2015; Mühlbacher & Kaczynski, 2016; Wahlster, Goetghebeur, Kriza, Niederländer, & Kolominsky-Rabas, 2015). MCDA is a methodology for evaluating alternatives based on individual preference, often against conflicting criteria, and combining them into one single appraisal. Due to its versatility, MCDA approaches are widely adopted in the fields of transportation, immigration, education, investment, environment, energy, defense and healthcare (Devlin & Sussex, 2011; Dodgson, Spackman, Pearman, & Phillips, 2009; Gregory et al., 2012; Mühlbacher & Kaczynski, 2016; Nutt, King, & Phillips, 2010; Wahlster et al., 2015).

The applications of MCDA in supplier selection are numerous and extensive. As one of the most popular MCDA method, Analytical Hierarchy Process (AHP) is widely adopted for supplier selection (De Felice, Deldoost, Faizollahi, & Petrillo, 2015; Prasad, Prasad, Rao, & Patro). In AHP, a complex master problem could be decomposed to plenty of sub-problems in several levels, in which way the unidirectional hierarchical relationships between levels are more understandable. Then pairwise comparison between alternatives are conducted to determine the importance of the criteria and preference over all alternatives. With the help of AHP, the evaluation of alternatives can be extended to multiple qualitative criteria along with satisfying the consistency of the decision making process. However, as the qualitative data are given by the decision makers based on experience, knowledge and judgment, the discrepancy among the decision makers, which would result in subjective influence in data, are not considered. Thus, the uncertainty and imprecise nature in the data are not dealt with, which may lead to low reliability and robustness of the result.

To develop a more accurate and reliable approach to evaluate the alternatives, TOPSIS (Technique for order preference by similarity to an ideal solution) was developed by Hwang et al. (Hwang, Lai, & Liu, 1993) based on the concept that the optimal solution should have the closest distance from the Positive Ideal Solution (PIS) and longest distance from the Negative Ideal Solution (NIS). The PIS and NIS are determined by the objective of the components of the variables, and the distances are usually measured by Euclidean Distance. Different from AHP, the input data are quantitative numbers so that the computation can be processed. Due to its high accuracy, the application of TOPSIS in supplier selection is also numerous. One of these studies (Shahroudi & Tonekaboni, 2012) proposed an application of TOPSIS in the supplier selection process in Iran Auto Supply Chain. In his research, both Numerical and Linguistic evaluation criteria are considered. To evaluate all the criteria simultaneously with quantitative data, the authors assigned numerical numbers (without consideration of fuzziness of data set) to each linguistic class, and generated the quantitative decision matrix. After the normalization and calculation of entropy measurement for the quantitative decision matrix, the weight of each criterion is determined, and TOPSIS is then adopted to measure the performance of each alternative supplier. Finally, the list of preference of the alternative suppliers are generated based on their ranking score and the optimal supplier is selected. In addition, to better evaluate the alternative suppliers, some of the researchers aggregated AHP with TOPSIS to develop a comprehensive decision-making framework (Bhutia & Phipon, 2012; Şahin & Yiğider, 2014). However, most of these methods are developed using crisp information, and the uncertainty, impreciseness and fuzziness nature of the information extracted from real world are not considered.

One of the conventional ways to represent linguistic evaluation in terms of a crisp number is Likert scale method, which is a standard psychometric scale and often used in research based on survey questionnaires to measure a particular response (Li, 2013; Liao, Xu, & Zeng, 2014). It is significantly different from possibility theory based linguistics term set (fuzzy sets in general) as Likert scale can only capture discrete measurement given by a respondent for an alternative in terms of a closed-form scaling or ordinal number. The experts participating in decision making process don't have other options but to choose from the given preferences, which doesn't necessarily comply with the exact responses (Hodge & Gillespie, 2003). Regardless of whether a Likert scale follows a closed-form or ordinal scale measurement method, it suffers from two phenomena while evaluating alternatives—information lost associated with closed-form scale and information distortion associated with ordinal scale (Russell & Bobko, 1992). Furthermore, adopting such Likert scale techniques in evaluating a set of alternatives against multiple evaluation criteria must assume that the information e.g., importance of decision makers and criteria are possible to represent in terms of crisp numbers. However, in most of the real-world decisions, information is generally incomplete, imprecise and vague in nature. The reasons behind this are (1) decision makers has to provide their judgmental evaluation under time pressure and lack of sufficient knowledge or data and (2) limited capacity for processing information, which altogether makes it challenging to estimate exact perceptual preferences with the help of crisp numerical value (Garg, 2016).

Therefore, in such a situation where making decision is subject to imprecise information entailing uncertainty, Likert scale-based decision is neither ideal nor reliable for strategic decisions. While crisp data is inadequate to model real life situations, intuitionistic fuzzy set (IFS) was introduced by Atanassov (Atanassov, 1986) and is adopted to the aggregated decision-making framework by Haldar et al. and Boran et al. (Boran, Genç, Kurt, & Akay, 2009; Haldar, Ray, Banerjee, & Ghosh, 2014). The intuitionistic fuzzy set entails both the truth-membership degree $t(x)$ and falsity membership degree $f(x)$ with $t(x), f(x) \in [0, 1]$ and $0 \leq t(x) + f(x) \leq 1$, and can better handle the incomplete information. Because IFS has the abilities to handle uncertainties in the data during decision making process, Kumar and Garg studied set pair analysis (SPA) based on IFS. They constructed connection number, a major component of SPA, and based on it, further extended the TOPSIS method to generate the ranking order of alternatives in Multi-Attribute Decision Making (MADM) approach under IFS environment (Garg, 2016). Later on, same group of authors developed some new distance measures based on the connection number of SPA and applied these in MADM framework under IFS (Garg & Kumar, 2018).

As in IFS environment, it is often difficult for an expert to exactly quantify his or her opinion as a number in interval [0, 1], it's more convenient to represent the preferences or opinions in terms of an interval rather than a crisp value, leading to the conception of interval-valued fuzzy set. Wang and Li (Guijun & Xiaoping, 1998) defined Interval-valued Fuzzy Sets (IVFS) and it has been widely applied in real-world problems. An IVFS is defined by a mapping an element of the universe of discourse to the closed interval in [0, 1]. IVFS includes two elements, which are the lower limit and upper limit of degree of membership. Thus, it enables an expert to provide his belief regarding a statement with an interval rather than a point value, and better represent the incompleteness and impreciseness of an information. Extending fuzzy set-theoretic operations to intervals canonically, the operation principles such as union, intersection etc. can be performed.

Interval-valued fuzzy set (IVF) and interval-valued intuitionistic fuzzy set (IVIFS) as an extension of fuzzy set theory is widely accepted due to their capability of handling impreciseness and uncertainty in the data (Atanassov, 1999; Garg, 2016). Unlike Likert scale, which lacks the sophistication of using compatibility function while measuring the decision maker's opinions (Liao et al., 2014), the IVF set utilize compatibility function e.g., triangular type membership function for triangular IVF set (Ashtiani, Haghighirad, Makui, & ali Montazer, 2009). It enables IVF set to assess an imprecise information and thus quantify inherent uncertainty in a systematic way by mathematically computing the compliance of a piece of information against a linguistic function. Leveraging the strength of IVIFS for solving Multiple Attribute Decision Making (MADM), Kumar and Garg further investigated the SPA under IVIFS environment. Using the connection numbers of the SPA, TOPSIS method was extended to IVFS and preferential ranking order was generated to select the best alternative in MADM setting (Kumar & Garg, 2018). Because the interval-valued fuzzy set theory can provide a more accurate modeling of uncertainty associated with incomplete information, Ashtiani et al. (Ashtiani et al., 2009) extend the application of IVFS in TOPSIS to solve Multi Criteria Decision Making problems. The algorithm of the interval-valued fuzzy TOPSIS is proposed and a numerical example considering linguistic criteria are presented to demonstrate the practicability of the model.

To characterize the indeterminacy associated with judgmental information more explicitly, the concept of neutrosophic set (NS) was introduced by Smarandache (Smarandache, 1998). To represent uncertainty, NS utilize three components, which are truth-membership degree, the indeterminacy-membership degree and the falsity membership degree, respectively. For example, in a situation when an expert is asked to provide his or her opinion regarding a particular statement, he or she may say that the possibility for that statement to be true is 0.4 and to be false is 0.7, and the possibility that he or she is not sure about the statement is 0.3. Such expert opinion under neutrosophic notation can be expressed as x (0.4, 0.3, 0.7). All these three membership degrees are independent and takes the value in the non-standard unit interval of $(0^-, 1^+)$, providing additional strength to represent uncertainty, imprecise, incomplete, and inconsistent information existing in real world. Thus, NS can successfully generalize the notion of indeterminacy from a philosophical point of view. To facilitate its strength for solving real-world scientific and engineering problem Wang et al. (Haibin, Smarandache, Zhang, & Sunderraman, 2010; Wang, Smarandache, Sunderraman, & Zhang, 2005) proposed the concept of *single valued neutrosophic set* (SVNS) as an instance of neutrosophic set (NS) and provided set-theoretic operators and properties of SVNS. With these advances, Garg and Nancy investigated SVNS theory with Multi-Attribute group Decision Making framework to simultaneously consider priority of attributes and uncertainty in linguistic term. They proposed prioritized weighted, ordered weighted and geometric aggregation operators to process linguistic single-valued neutrosophic numbers (Garg, 2018a). Later on, same group of authors introduced SVN prioritized murihead mean (SVNPMM) and SVN prioritized dual murihead mean (SVNPDMM) operators to solve MCDM problem under SVNS environment (Garg, 2018b). A hybrid aggregation operators considering arithmetic and geometric aggregation is proposed in a MCDM framework to deal with the preferences of attributes given in terms of single-valued and interval neutrosophic numbers (Garg, 2018c). As SVNS is more suitable to handle the uncertainty, imprecise, inconsistent and incomplete information existing in real world, Sahin and Yiğider (Şahin & Yiğider, 2014) introduced SVNS in TOPSIS to replace the crisp data in the decision matrix, and the results showed that the single valued neutrosophic TOPSIS can be preferable for dealing with incomplete, undetermined and inconsistent information in MCDA problems. However, only linguistic criteria were considered, while many of the essential and significant evaluation criteria may be expressed in numerical form in the process of supplier evaluation.

The existing research have shown promising potential of MCDA methods in supplier selection problems. However, there is still scope of extending the current methods to incorporate several aspects of real world supply chain resilience. The study done by Shahroudi and Tonekaboni (Shahroudi & Tonekaboni, 2012) presented the application of TOPSIS in supplier selection, where both linguistic and numerical criteria were considered. The works of Ashtiani et al. (Ashtiani et al., 2009) and Sahin and Yiğider (Şahin & Yiğider, 2014) showed the effectiveness of IVFS and SVNS within the TOPSIS process. However, to the best of our knowledge, there is no such method currently present that can evaluate the alternative suppliers from resilience perspective based on numerical and linguistic criteria simultaneously while considering the impreciseness and unreliable nature of the DRI provided by multiple decision makers. Focused on this relatively less studied problem, the originality of this paper lies in developing a comprehensive decision-making framework to help decision maker select *resilient suppliers.* The underlying idea of the proposed algorithm is to extend fuzzy based TOPSIS to a group decision making framework for evaluating alternative suppliers from resiliency perspectives. Although, prior research has separately explored TOPSIS with SVNS and IVFS, and group decision making approach, the integration of these two approaches particularly for *resilient supplier selection* problem is unique and carries the potential for planning better strategic sourcing decisions. Furthermore, in our proposed method, the reliability-based membership degree is adopted to bridge the gap in considering crisp data and the fuzzy set (SVNS and IVFS) on numerical criteria simultaneously to help decision maker evaluate the alternative set of suppliers from resilient point of view in a comprehensive and systematic way. In Table 1, we summarize the differences (advantages) of the proposed approach compared to other existing studies focusing on the application of MCDM approach in supplier selection problem.

We organize the rest of the paper as follows: in the Section 2, we introduce the methodology and present the preliminary concepts for designing the algorithm. Then in Section 3, we present the detailed computation process of the algorithm. In Section 4, we present an illustrative example of a resilient supplier selection problem following our algorithm. Finally, in Section 5 we conclude our paper with discussion on our findings and potential future research.

Table 1. Comparison Showing Advantages (Differences) of the Proposed Approach with Existing Studies

| Proposed decision-making approach | Existing studies |
|---|---|
| • We extend fuzzy based TOPSIS approach for simultaneous consideration of quantitative and linguistics evaluation criteria from resiliency perspective and systematic quantification of uncertainty associated with decision relevant information | • Existing study that considers both the quantitative and linguistic criteria did not demonstrate the formal quantification of uncertainty associated with linguistic information (Shahroudi & Tonekaboni, 2012). Only a single value was used corresponding to a linguistic class similar to Likert scale, which does not necessarily represent the uncertainty, and thus limiting the reliability of such decision-making approach. |
| • A decision-making framework where the opinion of multiple or a group of decision makers across all the quantitative and linguistic criteria can be integrated. This integration takes into account the importance of individual decision maker as well as criteria to generate an integrated decision matrix for SVNS and IVFS, separately.<br>• We process the quantitative criteria as crisp granular information and calculated their compliance value in response to each linguistic class. | • Prior studies that used TOPSIS along with the SVNS and IVFS, and multiple decision makers only considered linguistic criteria (Ashtiani et al., 2009; Şahin & Yiğider, 2014). However, no evidence was found on how to deal with uncertainty associated with quantitative information that entails crisp granular information resulting from the opinion of a group of decision makers. |
| • For the first time, we apply reliability modification technique in MCDM, where the objective is to consider the uncertainty and impreciseness resulted from the symmetric triangular membership functions. Such consideration of reliability index can improve the quality and reliability of MCDM approach, yet a comparative analysis is needed to be performed to assess the superiority of this addition. | • To the best of our knowledge, no prior research that studied MCDM for supplier selection problem did not modify the compliance or membership value using reliability index. |

## 2. Methodology

In the proposed MCDA model, we adapted two fuzzy-based approaches, IVFS and SVNS, to characterize the assessment given by decision makers as well as the weight of criteria and decision makers. Although these two approaches share some similarities; the primary difference is how the linguistic evaluations are represented by fuzzy numbers. Firstly, we computed the weight of criteria based on IVFS and SVNS theory and fuzzified the universe of discourse of numerical criteria according to the obtained weight. With the fuzzified frame of discernment, the membership function and associated membership degree for each linguistic class is calculated. After that, the derived membership degrees are modified with respect to reliability indices and normalized to be regarded as the weight of each class in a certain criterion for the supplier's performance measurement. Multiplied by the components of corresponding IVFS or SVNS of each linguistic class, the weighted average of each component is generated and the integrated IVFS or SVNS decision matrixes for numerical criteria are constructed. Then the obtained decision matrixes are coordinated with respect to the weight of criteria based on the algorithm for these two fuzzy approaches. Possessing the weighted decision matrixes, TOPSIS method is processed to compute the closeness coefficient (CC), which is deemed to be the ranking score to determine the list of preference of suppliers. To make it more understandable, we introduce some preliminaries of our methodology in the following subsections.

### 2.1. Multi Criteria Decision Analysis

Multi Criteria Decision Analysis (MCDA) provides a comprehensive decision analysis framework that could help the stakeholders balance the pros and cons of the alternatives in a multi-dimensional optimization problem, in which alternatives, evaluation criteria and decision makers are the essential variables. The analysis process of MCDA and the corresponding steps in our decision-making model is summarized as follows in the following Table 2 (Thokala et al., 2016):

Table 2. Framework of MCDA.

| | | |
|---|---|---|
| Step 1 | Defining the decision problem | Select optimal supplier with highest resilience over a group of alternative suppliers |
| Step 2 | Selecting and structuring criteria | Identify the evaluation criteria with respect to supplier resilience |
| Step 3 | Measuring performance | Gather data about the alternatives' performance on the criteria and summarize this in a decision matrix |
| Step 4 | Scoring alternatives | Evaluate the performance of the alternative suppliers based on the objective of the criteria |
| Step 5 | Weighting criteria and decision makers | Determine the weight of criteria and decision makers based on their importance |
| Step 6 | Calculating aggregate scores | Use the alternatives' scores on the criteria and the weights for the criteria and decision makers to get "total value" by which the alternatives are ranked with TOPSIS |

| Step 7 | Dealing with uncertainty | Perform Sensitivity analysis to understand the level of robustness of the MCDA results |
| Step 8 | Reporting and examination of findings | Interpret the MCDA outputs, including sensitivity analysis, to support decision making |

## 2.2. Technique for order preference by similarity to an ideal solution (TOPSIS)

TOPSIS is a decision-making technique wherein the alternatives are evaluated based on their distance to the ideal solution. The closer the distance of an alternative to the ideal solution, the higher a grade it would gain. In our research, Euclidian Distance is used to measure the performance of the alternatives and the function is described as below (Şahin & Yiğider, 2014):

$$s_i^+ = \sqrt{\sum_{j=1}^{n} \left\{ (a_{ij} - a_j^+)^2 + (b_{ij} - b_j^+)^2 + (c_{ij} - c_j^+)^2 \right\}} \quad i = 1, 2, \ldots, n \tag{2.1}$$

$$s_i^- = \sqrt{\sum_{j=1}^{n} \left\{ (a_{ij} - a_j^-)^2 + (b_{ij} - b_j^-)^2 + (c_{ij} - c_j^-)^2 \right\}} \quad i = 1, 2, \ldots, n \tag{2.2}$$

$$\tilde{\rho}_j = \frac{s^-}{s^+ + s^-}, \quad 0 \leq \tilde{\rho}_j \leq 1 \tag{2.3}$$

Where $s_i^+$ and $s_i^-$ are the positive and negative ideal solution respectively, $\tilde{\rho}_j$ is the closeness coefficient(CC), $a_{ij}, b_{ij}, c_{ij}$ are the component of the alternatives on criteria $j$ and $a_j^+, b_j^+, c_j^+$ are the corresponding components of the Positive Ideal Solution (PIS) and $a_j^-, b_j^-, c_j^-$ are the corresponding components of Negative Ideal Solution (NIS).

## 2.3. Interval Valued Fuzzy Set (IVFS)

An interval valued fuzzy set A defined on $(-1, +1)$ is given by (Haldar, Ray, Banerjee, & Ghosh, 2012):

$$A = \{(x, [\mu_A^L(x), \mu_A^U(x)]\}$$
$$\mu_A^L(x), \mu_A^U(x): X \to [0,1] \quad \forall x \in X, \mu_A^L(x) \leq \mu_A^U(x) \tag{2.4}$$
$$\mu_A(x) = [\mu_A^L(x), \mu_A^U(x)]$$
$$A = \{(x, \mu_A(x))\}, x \in (-\infty, +\infty)$$

; where $\mu_A^L(x)$ is the lower limit of degree of membership and $\mu_A^U(x)$ is the upper limit of degree of membership.

If the membership degree is expressed in triangular interval valued fuzzy numbers, it can be also demonstrated as:

$$x = [(x_1, x_1'); x_2; (x_3', x_3)] \tag{2.5}$$

; where $x_1$, $x_1'$, $x_2$, $x_3'$ and $x_3$ could be illustrated as shown in Figure 1:

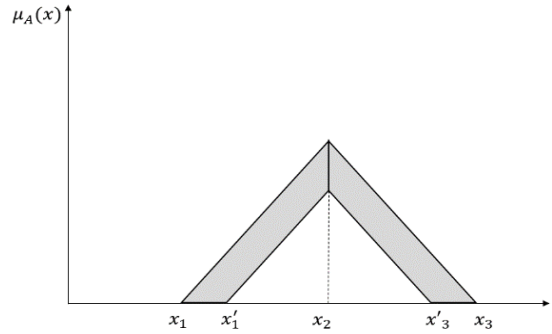

Figure 1. Illustration for the Triangular Interval Valued Fuzzy Number.

Table 3 shows the corresponding IVFS according to the linguistic terms:

Table 3. Linguistic Terms and Associated IVFS.

| Linguistic Terms | IVFS (a, a', b, c', c) | | | | |
|---|---|---|---|---|---|
| Weight Linguistic Terms in IVFS | | | | | |
| VUI | 0 | 0 | 0 | 0.15 | 0.15 |
| UI | 0 | 0.15 | 0.3 | 0.45 | 0.55 |
| M | 0.25 | 0.35 | 0.5 | 0.65 | 0.75 |
| I | 0.45 | 0.55 | 0.7 | 0.8 | 0.95 |
| VI | 0.55 | 0.75 | 0.9 | 0.95 | 1 |
| Performance Linguistic Terms in IVFS | | | | | |
| VB | 0 | 0 | 0 | 1 | 1.5 |
| B | 0 | 0.5 | 1 | 2.5 | 3.5 |
| MB | 0 | 1.5 | 3 | 4.5 | 5.5 |
| M | 1 | 2.5 | 4 | 5.5 | 6.5 |
| MG | 2.5 | 3.5 | 5 | 6.5 | 7.5 |
| G | 4.5 | 5.5 | 6 | 7 | 8.5 |
| VG | 5.5 | 6.5 | 7 | 8 | 9.5 |
| VVG | 7.5 | 8.5 | 9 | 9.5 | 10 |
| EG | 8.5 | 9.5 | 10 | 10 | 10 |

In Table 3, {VUI, UI, M, I, VI } is a set of linguistic weights for decision makers refers to Very Unimportant, Unimportant, Medium, Important and Very Important respectively, and {VB, B, MB, M, MG, G, VG, VVG, EG } is a set of linguistic weights for criteria refers to Very Bad, Bad, Medium Bad, Medium, Medium Good, Good, Very Good, Very Very Good and Extremely Good respectively.

For two IVFS $v = [(v_1, v_1'); v_2; (v_3', v_3)]$ and $u = [(u_1, u_1'); u_2; (u_3', u_3)]$, the algorithm for finding the compound IVFS is as follows:

$$v * u = [(v_1 * u_1, v_1^{'} * u_1^{'}); v_2 * u_2; (v_3^{'} * u_3^{'}, v_3 * u_3)] \qquad (2.6)$$

; where $* \in \{+, -, \times, \div\}$

Assume there are t decision makers, $V_k = [(v_{1k}, v_{1k}^{'}); v_{2k}; (v_{3k}^{'}, v_{3k})]$ refers to the weight of kth decision maker in the form of IVFS, $I_{jk} = [(u_{1jk}, u_{1jk}^{'}); u_{2jk}; (u_{3jk}^{'}, u_{3jk})]$ represent the weight of jth criteria given by kth decision maker in the form of IVFS, then the aggregated weight of jth criteria in the form of IVFS could be calculated based on (2.6) as follows:

$$I_j = [(\tfrac{1}{t}\Sigma_k v_{1k}u_{1jk}, \tfrac{1}{t}\Sigma_k v_{1k}^{'}u_{1jk}^{'}); \tfrac{1}{t}\Sigma_k v_{2k}u_{2jk}; (\tfrac{1}{t}\Sigma_k v_{3k}^{'}u_{3jk}^{'}, \tfrac{1}{t}\Sigma_k v_{3k}u_{3jk})] \qquad (2.7)$$

Similarly, if the performance measurement for ith alternative given by kth decision maker $P_{ik} = [(e_{1ik}, e_{1ik}^{'}); e_{2ik}; (e_{3ik}^{'}, e_{3ik})]$, then the aggregated performance measurement of ith alternative in the form of IVFS could be calculated as follows:

$$P_i = [(\tfrac{1}{t}\Sigma_k v_{1k}e_{1ik}^{'}, \tfrac{1}{t}\Sigma_k v_{1k}^{'}e_{1ik}^{'}); \tfrac{1}{t}\Sigma_k v_{2k}e_{2ik}; (\tfrac{1}{t}\Sigma_k v_{3k}^{'}e_{3ik}^{'}, \tfrac{1}{t}\Sigma_k v_{3k}e_{3ik})] \qquad (2.8)$$

### 2.4. Single Valued Neutrosophic Set (SVNS)

A single valued neutrosophic set (SVNS) can be defined as follows (Şahin & Yiğider, 2014):

Let X be a universe of discourse. A single valued neutrosophic set A over X is an object having the form

$$A = \{\langle x, u_A(x), r_A(x), v_A(x) \rangle : x \in X\} \qquad (2.9)$$

; where $u_A(x): X \to [0,1]$, $r_A(x): X \to [0,1]$ and $v_A(x): X \to [0,1]$ with $0 \leq u_A(x) + r_A(x) + v_A(x) \leq 3$ for all $x \in X$. The intervals $u_A(x)$, $r_A(x)$ and $v_A(x)$ denote the truth- membership degree, the indeterminacy-membership degree and the falsity membership degree of x to A, and can be simplified as a, b, c respectively. Table 4 shows the corresponding SVNS according to the linguistic terms:

Table 4. Linguistic Terms and Associated SVNS.

| Linguistic Terms | SVNS (a, b, c) | | |
|---|---|---|---|
| Weight Linguistic Terms in SVNS | | | |
| VI | 0.9 | 0.1 | 0.1 |
| I | 0.75 | 0.25 | 0.2 |
| M | 0.5 | 0.5 | 0.5 |
| UI | 0.35 | 0.75 | 0.8 |
| VUI | 0.1 | 0.9 | 0.9 |
| Performance Linguistic Terms in SVNS | | | |
| VB | 0.2 | 0.85 | 0.8 |
| B | 0.3 | 0.75 | 0.7 |
| MB | 0.4 | 0.65 | 0.6 |
| M | 0.5 | 0.5 | 0.5 |
| MG | 0.6 | 0.35 | 0.4 |
| G | 0.7 | 0.25 | 0.3 |
| VG | 0.8 | 0.15 | 0.2 |
| VVG | 0.9 | 0.1 | 0.1 |
| EG | 1 | 0 | 0 |

Assume the decision makers' group consists of $t$ participants, $A_k = (a_k, b_k, c_k)$ express the SVNS importance of the $k$th decision maker. Then the weight of $k$th decision maker can be calculated as follows:

$$\sigma_k = \frac{a_k + b_k(\frac{a_k}{a_k+c_k})}{\sum_{k=1}^{t} a_k + b_k(\frac{a_k}{a_k+c_k})}, \; \sigma_k \geq 0 \; and \; \sum_{k=1}^{t} \sigma_k = 1. \tag{2.10}$$

The aggregated SVNS decision matrix $D$ with respect to decision makers is defined by $D = \sum_{k=1}^{t} \delta_k D^k$, where

$$D = \begin{pmatrix} d_{11} & \cdots & d_{1j} \\ \vdots & \ddots & \vdots \\ d_{i1} & \cdots & d_{ij} \end{pmatrix} \tag{2.11}$$

And

$$d_{ij} = (u_{ij}, r_{ij}, v_{ij}) = (1 - \prod_{k=1}^{t}(1 - a_{ij}^{(k)})^{\sigma_k}, \; \prod_{k=1}^{t}(b_{ij}^{(k)})^{\sigma_k}, \prod_{k=1}^{t}(c_{ij}^{(k)})^{\sigma_k}) \tag{2.12}$$

Let $w_j^{(k)} = (a_j^{(k)}, b_j^{(k)}, c_j^{(k)})$ be an SVN number expressing the importance of criteria j (j = 1, 2, ...) by the kth decision maker. The SVNS describing the weight of the jth criteria can be calculated using the method proposed by:

$$w_j = (1 - \prod_{k=1}^{t}(1 - a_{jk})^{\sigma_k}, \; \prod_{k=1}^{t} b_{jk}^{\sigma_k}, \; \prod_{k=1}^{t} c_{jk}^{\sigma_k}) \tag{2.13}$$

While the weight vector of all criteria is presented as:

$$W = (w_1, w_2, \ldots, w_j) \tag{2.14}$$

Then the aggregated weighted SVNS decision matrix can be obtained as:

$$D^* = D \otimes W \tag{2.15}$$

Based on the product algorithm of SVNS:

$$A_1 \otimes A_2 = (a_1 a_2, \; b_1 + b_2 - b_1 b_2, c_1 + c_2 - c_1 c_2) \tag{2.16}$$

The algorithm of SVNS is totally different to that of IVFS based on (2.6-2.8), so we adopt these two fuzzy approaches to demonstrate the effectiveness and robustness of our decision-making model.

## 2.5. Membership Function

The definition of membership function was first introduced by Zadeh (Zadeh, 1965), where the membership functions were used to operate on the domain of all possible values. In fuzzy logic, membership degree represents the truth value of a certain proposition. Different from the concept of probability, truth value represents membership in vaguely defined sets. For any set X, the membership degree of an element x of X in fuzzy set A is denoted as $\mu_A(x)$, which quantifies the grade of membership of the element x to the fuzzy set A.

Ullah et al. (Ullah & Noor-E-Alam, 2018) presented a method to fuzzify the universe of discourse of the numerical criteria and formulate the membership function to calculate the membership degree with range data as shown in Figure 2:

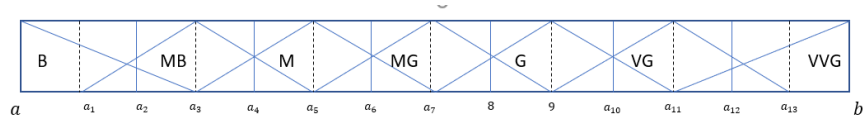

Figure 2. Frame of Discernment.

For the frame of discernment shown in the above figure, the membership functions of the different classes (B, MB, M, MG, G, VG, VVG) are calculated as follows:

$$m_B = max\left(0, \frac{a_3 - x}{a_3 - a}\right)$$
$$m_{MB} = \max\left(0, \min(\frac{x-a_1}{a_3-a_1}, \frac{a_5-x}{a_5-a_3})\right) \quad (2.17)$$
$$m_{VVG} = \max\left(0, \frac{x - a_{11}}{b - a_{11}}\right)$$

; where $a_i = a + \frac{b-a}{2\times 7}$, $i = 1, 2, \dots, (2 \times 7 - 1)$, and 7 is the number of class in this frame of discernment. The membership functions are assumed to be triangular and symmetric. The membership function for each class depends on the frame of discernment of the criteria.

## 2.6. Reliability-based Membership Function

As the membership functions are assumed triangular and symmetric, the uncertainty and impreciseness of the functions need to be taken into consideration. Jiang et al (Jiang, Zhuang, & Xie, 2017) proposed a reliability-based membership function to deal with uncertainty of information and the reliability of information sources. The reliability of the membership functions is measured by the static reliability index and dynamic reliability index, which are defined by the similarity among classes and the risk distance between the test samples and the overlapping area among classes, respectively. The comprehensive reliability is computed by the product of the two index and the reliability-based membership function are fused using Dempster's combination rule (Dempster, 1967; Shafer, 1976). The numerical examples provided by Jiang et al. verified the effectiveness of the reliability-modified functions.

The static reliability index is measured by the overlapped area between two adjacent classes. In Figure 3, the shaded area is the overlapped region between classes M and MG.

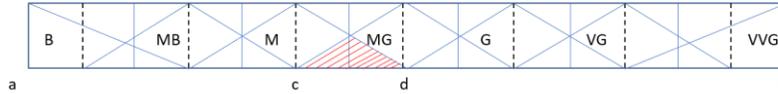

Figure 3. Illustration of Static Reliability Index.

The larger the overlapped area between classes M and MG, it is more likely that an input data is wrongly recognized to fall in M class while it actually belongs to MG class. The similarity between classes M and MG $sim_{M,MG}$ in a certain criterion can be described as (Jiang et al., 2017):

$$sim_{M,MG} = \frac{\int_c^d \min_{c \leq x \leq d}(m_M(x), m_{MG}(x))dx}{\int m_M(x) + m_{MG}(x) - \int_c^d \min_{c \leq x \leq d}(m_M(x), m_{MG}(x))dx} \qquad (2.18)$$

The static reliability index $R^s$ then can be calculated as:

$$R^s = \sum_{i<l}(1 - sim_{il}) \qquad (2.19)$$

While $i$ and $l$ are the adjacent classes in the same universe of discourse in one criterion.

The dynamic reliability index is measured with a set of test sample and the risk distance between every peak of overlap and the test value.

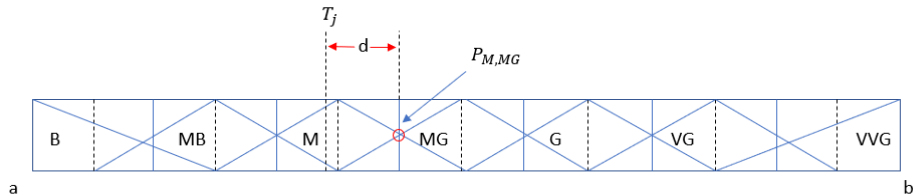

Figure 4. Illustration of Dynamic Reliability Index.

If $P_{M,MG}$ is the peak of the overlapping area between classes M and MG in Figure 4, and $T_j$ is the test sample for this criterion $C_j$, the distance $d$ between $T_j$ and $P_{M,MG}$ represents the risk distance that related to the uncertainty of the test sample.

The risk distance can be formulated as (Jiang et al., 2017):

$$d_{M,MG} = \frac{|T_j - P_{M,MG}|}{D} \tag{2.20}$$

; where D is the range of the universe of discourse of $C_j$, which is $(a - b)$.

After calculating all the risk distance, the total dynamic reliability index can be determined as:

$$R^d = e^{\sum_2^n d_{(l-1)l}} \tag{2.21}$$

Then the Comprehensive reliability index can be defined as:

$$R = R^s \times R^d \tag{2.22}$$

After the normalization:

$$R^* = \frac{R}{\max(R)} \tag{2.23}$$

The reliability-based membership degree can be calculated as:

$$m_l^R = R^* \times m_l \tag{2.24}$$

; where $l$ is a class in a universe of discourse.

### 2.7. Evaluation Criteria

The evaluation criteria are divided into two groups: Resiliency Criteria and Critical Criteria (Haldar et al., 2012). In the following subsections we describe the criteria considered in each group and a summary of those criteria is presented in Table 5.

Table 5. Description of Evaluation Criteria.

|  | Criteria | Description |
|---|---|---|
| Resiliency Criteria | Supply chain density | The quantity and geographical spacing of nodes within a supply chain |
|  | Supply chain complexity | The number of nodes in a supply chain and the interconnections between those nodes |
|  | Responsiveness | The response speed of the supplier to market demand |
|  | Number of critical nodes in a Supply chain | Node criticality is the relative importance of a given node or set of nodes within a supply chain |
| Critical Criteria | Re-engineering | The corrective procedure for the incorporation of any engineering design change within the product |
|  | Buffer capacity | The level of extra stock that is maintained to mitigate the risk of stock-outs due to uncertainties in supply and demand. |
|  | Supplier's resource flexibility | The different logistics strategies which can be adopted either to release a product to a market or to procure a component from a supplier |
|  | Lead time | The delay between the initiation and execution of a process |

**2.7.1 Resiliency criteria**

The term resilience extends the well-established concept of pre-disaster mitigation, which focuses on improving the performance of critical systems or infrastructures to minimize losses resulting from a disaster. The concept of resilience further emphasizes the enhancement in systems flexibility and performance both during and after a disaster. In line with it, a resilient supplier or a resilient supply chain is characterized by the adaptive capability in an effort to (i) respond to a disruption by minimizing its possibility of occurrence, (ii) reduce the negative consequences of a disruption once it strikes, and (iii) ensure faster recovery of overall supply chain to normal operating state. We identify the following four criteria that are considered as resilience measure during the evaluation of a resilient suppliers.

*2.7.1.1 Supply chain density*

The quantity and geographical distance of nodes in the supply chain is defined as supply chain density. A node in supplier's network is a physical connection point where several other sub-orders are joined towards a fulfillment of an ordered quantity. Close cluster of a large number of such sub-assembly nodes within a particular geographical region leads to a high-density supplier network. Although, clustering several nodes in one location is a viable approach for an efficient supply chain, the prevalence of high density nodes is susceptible to disruption given the high risk associated with these nodes. Thus, to increase resilience, a supplier needs to redistribute these high density nodes, potentially distributing the risk to multiple locations.

*2.7.1.2 Supply chain complexity*

Complexity of a supplier network is defined as the number of nodes and interconnections among these nodes. This attributes usually introduces inflexibility and inefficiency in a supply chain, indicating high potential for disruptions to affect larger areas of the supply chain. However, contracting to a new node (e.g., secondary supplier) as back-up removes the uncertainty in terms of ensuring the timely supply of a critical item, thus acting as buffer to increase overall resilience of a supplier. Therefore, higher complexity relates to relatively higher redundancy, yet eventually implies the higher resilience for suppliers.

*2.7.1.3 Responsiveness*

The speed with which a supplier responds to changing order placed by the company—mainly driven by volatile customer demand is termed as responsiveness. It is crucial for a strategic manufacturer in terms of gaining competitive advantage in the market. Furthermore, being responsive to fluctuating customer demand, at first place, depends on supplier's ability to meet orders with higher variety within shorter time. Thus, decision makers will opt for suppliers having quicker response if there is bottlenecks in supply side or uncertainty in the customer demand, increasing resilience in overall supply chain.

*2.7.1.4 Number of critical nodes in a supply chain*

Criticality of a node or set of nodes in a supply chain is the relative importance of a node (set of nodes) characterized by the presence of hubs, which play a significant role for the connectivity of entire supply chain. And, of course the failure of such hubs has the higher risk of damage resulting from disruption than that of non-critical nodes. Replacing these critical hubs (or critical nodes) results into ta supply chain that entails a series of unconnected subnets, ensuring less vulnerability to disruption, and eventually leads to higher resilience.

**2.7.2 Critical criteria**

Following four criteria are considered as critical criteria from manufacturer's perspective. These features of the suppliers are essential to respond to the change of internal business strategies or technological advancement in product development, which are critical for the overall productivity and profitability of a manufacturing company. Such changes often cause uncertainties in the demand of advanced products as well as supply quantities, which requires the suppliers to be more resilient. Furthermore, these critical attributes of the suppliers also become essential for the manufactures during disruptions.

*2.7.2.1 Re-engineering*

Re-engineering refers to reworking on existing products for some minor corrections and (or) incorporating certain design changes. It may also be defined as the reverse engineering procedure that allows for modification of existing product after identifying its built-in features. Re-engineering enhance product reliability or maintainability to fulfill emerging need of customer with minimal effort and cost. The importance of re-engineering further lies in its potential to improve functionality of an existing product while leveraging the advantage of newly introduced technologies, yet without making major alteration in the current product. It has been proven effective to face global competition resulting from the rapid enhancement of technologies, which has direct influence on the change of customer preference. Thus, to comply with this and to ensure resilience, decision makers should check the supplier's re-engineering ability.

*2.7.2.2 Buffer capacity*

Buffer—a widely used terminology in logistics describes the level of safety stock that is maintained to minimize the loss of anticipated stock-outs of raw materials resulting from the uncertainty of supply and demand. Having sufficient level of buffer stock helps supplier and in turn manufacturer to be resilient in the face of instantaneous needs originated from uncertain demand–supply combination and lead time for products.

*2.7.2.3 Supplier's resource flexibility*

This attribute of a supplier demonstrates the ability to have resource back-up to adopt different strategies needed to deliver a product or procure required materials trough alternative channel. This enables a supplier to respond to mitigate the negative impact of disruption by ensuring uninterrupted flow of supply of raw materials to the manufacturer.

*2.7.2.4 Lead time*

Lead time is defined as the time required for a supplier to execute or fulfill an order earlier placed by the manufacturer. Shorter lead time is preferred as it helps manufacturer to be on schedule in terms of production and fulfilling customer orders. Thus, suppliers with shorter lead time can mitigate the aftermath of disruption by enabling manufacturer to restore its production capacity to normal operating condition within less time.

## 3. Computation Process

To better illustrate the detailed steps of our proposed decision-making algorithm, we provide a comprehensive flow that describes the computation process. The computation process includes both the IVFS and SVNS approaches and is summarized as follows:

*Step 1: Categorize the criteria into different group.*

Firstly, we determined the feature and objective of each criterion to prepare for further computation. In Table 6, we present the criteria for resilient supplier selection and the objectives in regard to each of them.

Table 6. Categorization of Criteria.

| Categorization | | Criteria | Obj. |
|---|---|---|---|
| Numerical | $C_1$ | Number of critical nodes in a Supply chain | Min |
| | $C_2$ | Buffer capacity | Max |
| | $C_3$ | Lead time | Min |
| Linguistic | $C_4$ | Supply chain density | Min |
| | $C_5$ | Supply chain complexity | Max |
| | $C_6$ | Responsiveness | Max |
| | $C_7$ | Re-engineering | Max |
| | $C_8$ | Supplier's resource flexibility | Max |

*Step 2: Calculate the weight of each decision maker and criterion.*

To determine the frame of discernment of the numerical criteria, we need to obtain their weight in advance. Suppose we have i alternative suppliers $S_i$, j evaluation criteria $C_j$ and k decision makers $DM_k$. Assume $D_k$ refers to the linguistic weight of $DM_k$, $L_{jk}$ refers to the importance of criteria j given by $DM_k$, based on Table 3 and equation (2.10) and (2.13), the weight of each DM and criterion in SVNS approach could be calculated as $w_{cj}$, as shown in Table 7.

Table 7. Calculated SVNS for Weight of Criteria.

| Criteria | Weight | | |
|---|---|---|---|
| | a | b | c |
| $C_1$ | $w_{a1}$ | $w_{b1}$ | $w_{c1}$ |
| $C_2$ | $w_{a2}$ | $w_{b2}$ | $w_{c2}$ |
| … | … | … | … |
| $C_j$ | $w_{aj}$ | $w_{bj}$ | $w_{cj}$ |

For IVFS approach, based on (2.7), we just need to transfer the linguistic data to IVFS data for further computation.

*Step 3: For the numerical criteria, fuzzify their universe of discourse based on their importance and determine the membership function of the classes.*

After we get the weight of the numerical criteria, the number of classes and frame of discernment could be determined. Based on the obtained weight, the number of classes in each corresponding criterion can be assigned. The higher the weight of criteria, the more number of classes are divided.

In the fuzzification process, the span of the universe of discourse is generated by the minimum and maximum of all the crisp data for a certain criterion including all the alternative suppliers. For example, if the minimum and maximum of crisp data in $C_j$ are $a_0$ and $a_{18}$ concerning all the suppliers in $C_j$ then the span of universe of discourse of $C_j$ is $(a_0, a_{18})$. Assume we decide to assign 9 classes to $C_j$ (as shown in Figure 5) based on the weight of $C_j$, then the universe of discourse underlying the regular scheme of $C_j$ can be fuzzified as follows (Ullah & Noor-E-Alam, 2018):

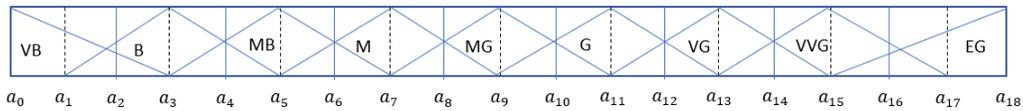

Figure 5. Illustration of Fuzzification.

$$a_i = a_0 + \frac{a_{18}-a_0}{18}, \ i = 1, 2, \dots, 18 \tag{3.1}$$

The membership function of each class $m_F$ in Figure 5 can be formulated as (Ullah & Noor-E-Alam, 2018):

$$m_{VB} = max\left(0, \frac{a_3-x}{a_3-a_0}\right)$$

$$m_B = \max\left(0, \min\left(\frac{x-a_1}{a_3-a_1}, \frac{a_5-x}{a_5-a_3}\right)\right) \qquad (3.2)$$

$$m_{EG} = \max\left(0, \frac{x-a_{15}}{a_{18}-a_{15}}\right)$$

*Step 4: Transfer the crisp data in the numerical criteria to range data and calculate membership degree.*

Firstly, we put the crisp data in an ascending order. Take numerical criteria $C_j$ for example, if we have $S_i$ as the $i^{th}$ supplier, $N_{ijk}$ as the crisp data for supplier $i$ on criteria $j$ given by decision maker $k$, $N_{ijn}$ and $N_{ijm}$ are the minimum and maximum values for $S_i$ on $C_j$, a representative crisp data set for this criteria is shown in Table 8.

Table 8. Crisp Data of $C_j$.

| Supplier | $C_j$ | | | | Min | Max |
|---|---|---|---|---|---|---|
| | $DM_1$ | $DM_2$ | … | $DM_k$ | | |
| $S_1$ | $N_{1j1}$ | $N_{1j2}$ | … | $N_{1jk}$ | $N_{1jn}$ | $N_{1jm}$ |
| $S_2$ | $N_{2j1}$ | $N_{2j2}$ | … | $N_{2jk}$ | … | … |
| … | … | … | … | … | … | … |
| $S_i$ | $N_{ij1}$ | $N_{ij2}$ | … | $N_{ijk}$ | … | … |

Assume that $N_{1j1} < N_{1j2} < … < N_{1jk}$ for $S_1$, the range value of $S_1$ on $C_j$ could be described as $r_{1j}$ ($N_{111}$, $N_{11k}$). For $S_1$, there are k crisp data given by decision makers through $DM_1$ to $DM_k$. These k data consist the data sample for $S_1$ concerning $C_j$. Regarding these k crisp data as a data set for $S_1$, the range of this data set is from the minimum to the maximum, which are $N_{111}$ and $N_{11k}$. So, the range value for $S_1$ is ($N_{1j1}$, $N_{1jk}$). This way, all the k crisp data given by the decision makers are aggregated in the range value and all their contribution are included in this range value. In this process, we don't consider the weight of decision makers. After that, the decision makers are no longer involved in the computation process of Numerical Criteria because their contribution has been presented in the range value, which is considered as the fundamental input through the whole model.

After we get the range value, we can calculate membership degree for each class according to the membership function (Ullah & Noor-E-Alam, 2018):

$$m_{ij} = \frac{\int_{x \in R} m_F(x)dx}{\|\acute{R}\|} \qquad (3.3)$$

; where R refers to the span of the criteria and $\|\acute{R}\|$ refers to the largest segment of R that belongs to the support $m_F$.

This way, the membership degree of $S_i$ on $C_j$ at linguistic class $l$ could be calculated as $M_{ijl}, l = 1, 2, 3, \ldots, 9$.

*Step 5: Calculate the normalized reliability-based membership degree.*

According to function (2.18-2.24), comprehensive reliability indexes for $C_j$ could be generated as $Rc_j$. Then based on (2.25), the original membership degree can be modified. To get the normalized weight of each class and make them add up to 1, we should normalize the membership value as:

$$n_{ijl} = \frac{m_{ijl}}{\sum_{l=1}^{9} m_{ijl}} \quad (3.4)$$

Then the original membership degree $M_{ijl}$ can be normalized as $A_{ijl}$, Where $\sum_{l=1}^{9} A_{ijl} = 1$, $l = 1, 2, 3, \ldots, 9$

*Step 6: Generate the integrated SVNS and IVFS for every alternative supplier concerning each criterion.*

To get SVNS for the numerical criteria, we integrated SVNS and the membership degree we calculated above to generate the integrated SVNS for the numerical criteria. In this integration process, the membership degree is regarded as the weight of each linguistic class for every alternative on the criteria. Then the membership degree is multiplied by the components of the corresponding SVNS presented in Table 4 to obtain the integrated SVNS. The components of the integrated SVNS for each $S_i$ on $C_j$ could be calculated as:

$$a_{ij} = \sum_l A_{ijl} \, a_l, l = 1, 2, \ldots, 9 \quad (3.5)$$

$$b_{ij} = \sum_l A_{ijl} \, b_l, l = 1, 2, \ldots, 9 \quad (3.6)$$

$$c_{ij} = \sum_l A_{ijl} \, c_l, l = 1, 2, \ldots, 9 \quad (3.7)$$

; where $a_l$, $b_l$ and $c_l$ are the corresponding components of class $l$ in Table 4.

The integrated SVNS for $S_i$ on $C_j$ can be described as $(a_{ij}, b_{ij}, c_{ij})$, Where $j = 1, 2, 3$. Similarly, based on Table 3, the components of the integrated IVFS for each $S_i$ on $C_j$ could also be calculated based on following functions:

$$a_{ij} = \sum_l A_{ijl} \, a_l, l = 1, 2, \ldots, 9 \quad (3.8)$$

$$a'_{ij} = \sum_l A_{ijl} \, a'_l, l = 1, 2, \ldots, 9 \quad (3.9)$$

$$b_{ij} = \sum_l A_{ijl} \, b_l, l = 1, 2, \ldots, 9 \quad (3.10)$$

$$c'_{ij} = \sum_l A_{ijl}\, c'_l, l = 1, 2, \ldots, 9 \tag{3.11}$$

$$c_{ij} = \sum_l A_{ijl}\, c_l, l = 1, 2, \ldots, 9 \tag{3.12}$$

; where $a_l$, $a'_l$, $b_l$, $c'_l$ and $c_l$ are the corresponding components of class $l$ in Table 3.

*Step 7: Construction of aggregated decision matrix with respect to decision makers for linguistic criteria.*

If $L_{ijk}$ refers to the linguistic data of $S_i$ on $C_j$ given by $DM_k$, based on (2.11) and (2.12) and Table 4, the aggregated SVNS with respect to decision makers can be obtained as $(a_{ij}, b_{ij}, c_{ij})$, where $j = 4, 5, \ldots, 8$.

Similarly, based on (2.8) and Table 3, the aggregated IVFS with respect to decision makers can be obtained as well.

*Step 8: Aggregate the numerical criteria matrix and the linguistic criteria matrix.*

As we have got all the SVNS for both numerical and linguistic criteria, we are able to build up a complete SVNS matrix. Similarly, the decision matrix as IVFS could also be generated.

*Step 9: Construction of aggregated weighted decision matrix with respect to criteria*

By using the weight of criteria matrix Table 7 and the complete SVNS decision matrix, the aggregated weighted SVNS decision matrix can be obtained based on (2.15) and (2.16), which is presented in Table 9.

Table 9. SVNS Decision Matrix.

| Supplier | Complete SVNS Decision Matrix | | |
|---|---|---|---|
| | SVNS | | |
| | $C_1$ | … | $C_j$ |
| $S_1$ | $(a_{11}, b_{11}, c_{11})$ | … | $(a_{1j}, b_{1j}, c_{1j})$ |
| $S_2$ | $(a_{21}, b_{21}, c_{21})$ | … | $(a_{2j}, b_{2j}, c_{2j})$ |
| … | … | … | … |
| $S_i$ | $(a_{i1}, b_{i1}, c_{i1})$ | … | $(a_{ij}, b_{ij}, c_{ij})$ |
| Aggregated Weighted SVNS Decision Matrix | | | |
| $S_1$ | $(a_{11}, b_{11}, c_{11})^*$ | … | $(a_{1j}, b_{1j}, c_{1j})^*$ |
| $S_2$ | $(a_{21}, b_{21}, c_{21})^*$ | … | $(a_{2j}, b_{2j}, c_{2j})^*$ |
| … | … | … | … |
| $S_i$ | $(a_{i1}, b_{i1}, c_{i1})^*$ | … | $(a_{ij}, b_{ij}, c_{ij})^*$ |

; where $j = 1, 2, \ldots, 8$

For the IVFS approach, we must normalize the decision matrix in advance. Given $x_{ij} = [\left(a_{ij}, a'_{ij}\right); b_{ij}; \left(c'_{ij}, c_{ij}\right)]$, based on the following functions (Ashtiani et al., 2009)

$$r_{ij} = \left[\left(\frac{a_{ij}}{c_j^+}, \frac{a_{ij}'}{c_j^+}\right); \frac{b_{ij}}{c_j^+}; \left(\frac{c_{ij}'}{c_j^+}, \frac{c_{ij}}{c_j^+}\right)\right], \quad \forall j \in G_1 \qquad (3.13)$$

$$r_{ij} = \left[\left(\frac{a_j^-}{a_{ij}}, \frac{a_j^-}{a_{ij}'}\right); \frac{a_j^-}{b_{ij}}; \left(\frac{a_j^-}{c_{ij}'}, \frac{a_j^-}{c_{ij}}\right)\right], \quad \forall j \in G_2 \qquad (3.14)$$

$$c_j^+ = \max_i c_{ij}, \quad \forall j \in G_1 \qquad (3.15)$$

$$a_j^- = \min_i a_{ij}', \quad \forall j \in G_2 \qquad (3.16)$$

; where $G_1 = \{C_2, C_5, C_6, C_7, C_8\}$, $G_2 = \{C_1, C_3, C_4\}$ based on Table 6.

Then, the aggregated weighted IVFS decision matrix can be obtained based on (2.8).

*Step 11: Determine positive-ideal solution and negative-ideal solution*

According to SVNS theory and the principle of classical TOPSIS method, SVNS-PIS and SVNS-NIS can be defined as below (Şahin & Yiğider, 2014):

$$\rho^+ = (a_j^+, b_j^+, c_j^+) \qquad (3.17)$$

$$\rho^- = (a_j^-, b_j^-, c_j^-) \qquad (3.18)$$

; where $\rho^+$ is the PIS and $\rho^-$ is the NIS and

$$a_j^+ = \begin{pmatrix} \max_i a_j, & if\ j \in G_1 \\ \min_i a_j, & if\ j \in G_2 \end{pmatrix} \qquad a_j^- = \begin{pmatrix} \max_i a_j, & if\ j \in G_2 \\ \min_i a_j, & if\ j \in G_1 \end{pmatrix}$$

$$b_j^+ = \begin{pmatrix} \max_i b_j, & if\ j \in G_2 \\ \min_i b_j, & if\ j \in G_1 \end{pmatrix} \qquad b_j^- = \begin{pmatrix} \max_i a_j, & if\ j \in G_1 \\ \min_i a_j, & if\ j \in G_2 \end{pmatrix}$$

$$c_j^+ = \begin{pmatrix} \max_i c_j, & if\ j \in G_2 \\ \min_i c_j, & if\ j \in G_1 \end{pmatrix} \qquad c_j^- = \begin{pmatrix} \max_i a_j, & if\ j \in G_1 \\ \min_i a_j, & if\ j \in G_2 \end{pmatrix}$$

For IVFS approach, however, the IVFS-PIS and IVFS-NIS could be defined as (Ashtiani et al., 2009):

$$\rho^+ = [(1,1); 1; (1,1)] \qquad (3.17)$$

$$\rho^- = [(0,0); 0; (0,0)] \qquad (3.18)$$

*Step 12: Calculate the Euclidian distance measures from SVN positive-ideal solution and SVN negative-ideal solution, the closeness coefficient (CC) and rank the alternatives*

Based on function (2.1), (2.2), and (2.3), the closeness coefficient (CC) of the alternatives are obtained and the list of preference are generated according to descending order for SVNS approach, meaning that higher the CC, the higher will be the ranking for a particular supplier. Ranking is given as an ascending order starting from 1 for a supplier with highest CC, and follows chronological order for rest of the suppliers

For IVFS approach, the Euclidian Distance could be calculated based on the following functions (Ashtiani et al., 2009):

$$D_{i1}^+ = \sum_j \sqrt{\frac{1}{3}[(a_{ij} - 1)^2 + (b_{ij} - 1)^2 + (b'_{ij} - 1)^2]} \tag{3.19}$$

$$D_{i2}^+ = \sum_j \sqrt{\frac{1}{3}[(a'_{ij} - 1)^2 + (b'_{ij} - 1)^2 + (b_{ij} - 1)^2]} \tag{3.20}$$

$$D_{i1}^- = \sum_j \sqrt{\frac{1}{3}[(a_{ij} - 0)^2 + (b_{ij} - 0)^2 + (b'_{ij} - 0)^2]} \tag{3.21}$$

$$D_{i2}^- = \sum_j \sqrt{\frac{1}{3}[(a'_{ij} - 0)^2 + (b'_{ij} - 0)^2 + (b_{ij} - 0)^2]} \tag{3.22}$$

The Closeness Coefficient for IVFS could be obtained by:

$$RC_1 = \frac{D_{i2}^-}{D_{i2}^+ + D_{i2}^-}, \quad RC_2 = \frac{D_{i1}^-}{D_{i1}^+ + D_{i1}^-} \tag{3.23}$$

$$RC_i^* = \frac{RC_1 + RC_2}{2} \tag{3.24}$$

The following flow diagram in Figure 6 shows the steps of our MCDA algorithm:

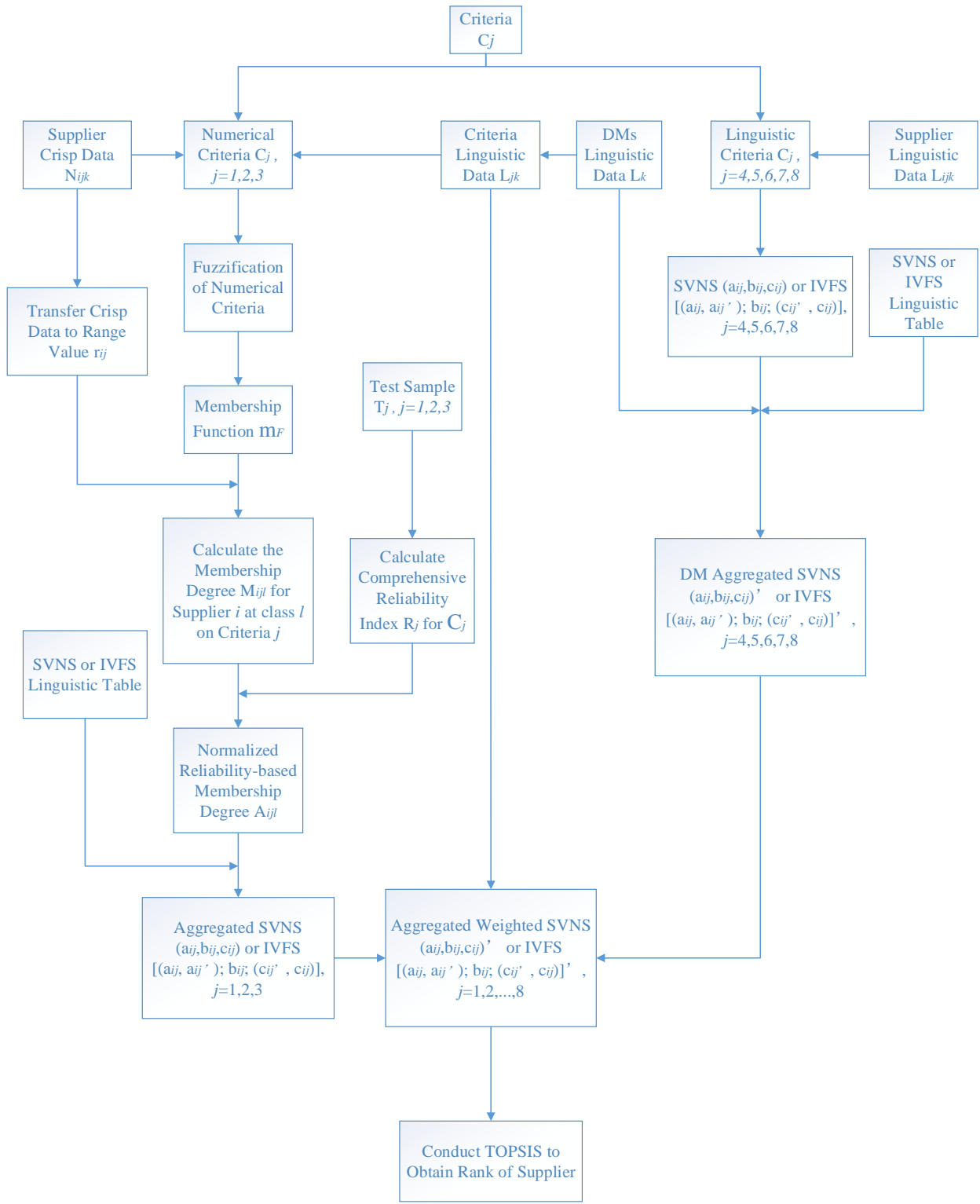

Figure 6. Framework of the Proposed MCDA algorithm.

## 4. Numerical Example

Since there is no existing data set available in the literature, and also obtaining supplier related data directly from the company entails confidentiality issue, to test the viability of our proposed method we generated a random data. To generate this data set, we assumed that that 10 decision makers (DM) evaluate 8 alternative suppliers with respect to 8 performance evaluation criteria (see section 2.7).

### 4.1 Result Analysis

The preferential ranking scores of the several alternative suppliers generated by the proposed algorithm using SVNS and IVFS approaches are presented in Table 10. Lowest ranking score represents higher priority (rank) for a particular supplier. A comparative analysis of the ranking scores given by these two approaches is presented in Figure 7. We found that the final ranks of the alternative suppliers are almost the same in both approaches, which show the robustness of the proposed decision-making algorithm. Apart from the slightly different ranking of suppliers 2 and 8, the final ranking of other suppliers were the same in both approaches.

Table 10. Ranking of the Suppliers.

| SVNS Approach | | | | |
|---|---|---|---|---|
| Supplier | d+ | d- | cc | Ranking Score |
| $S_1$ | 0.447 | 0.52 | 0.538 | 3 |
| $S_2$ | 0.656 | 0.409 | 0.384 | 8 |
| $S_3$ | 0.364 | 0.783 | 0.683 | 1 |
| $S_4$ | 0.639 | 0.476 | 0.427 | 6 |
| $S_5$ | 0.565 | 0.607 | 0.518 | 4 |
| $S_6$ | 0.644 | 0.512 | 0.443 | 5 |
| $S_7$ | 0.346 | 0.675 | 0.661 | 2 |
| $S_8$ | 0.617 | 0.428 | 0.409 | 7 |

| IVFS Approach | | | | | | | |
|---|---|---|---|---|---|---|---|
| Supplier | $d_1^+$ | $d_2^+$ | $d_1^-$ | $d_2^-$ | $RC_1$ | $RC_2$ | RC | Ranking Score |
| $S_1$ | 3.5 | 4.3 | 4.3 | 5 | 0.5 | 0.4 | 0.456 | 3 |
| $S_2$ | 5.3 | 4.7 | 3.1 | 3.9 | 0.5 | 0.4 | 0.409 | 7 |
| $S_3$ | 4.2 | 3.7 | 5.4 | 4.3 | 0.6 | 0.5 | 0.546 | 1 |
| $S_4$ | 3.4 | 4.5 | 4.2 | 5.1 | 0.5 | 0.4 | 0.442 | 6 |
| $S_5$ | 5.2 | 8.2 | 3.5 | 8.2 | 0.5 | 0.4 | 0.451 | 4 |
| $S_6$ | 5.2 | 5.2 | 3.4 | 5.4 | 0.5 | 0.4 | 0.45 | 5 |
| $S_7$ | 4.5 | 3.9 | 4 | 5.2 | 0.6 | 0.5 | 0.521 | 2 |
| $S_8$ | 5.5 | 4.9 | 2.9 | 3.6 | 0.4 | 0.3 | 0.385 | 8 |

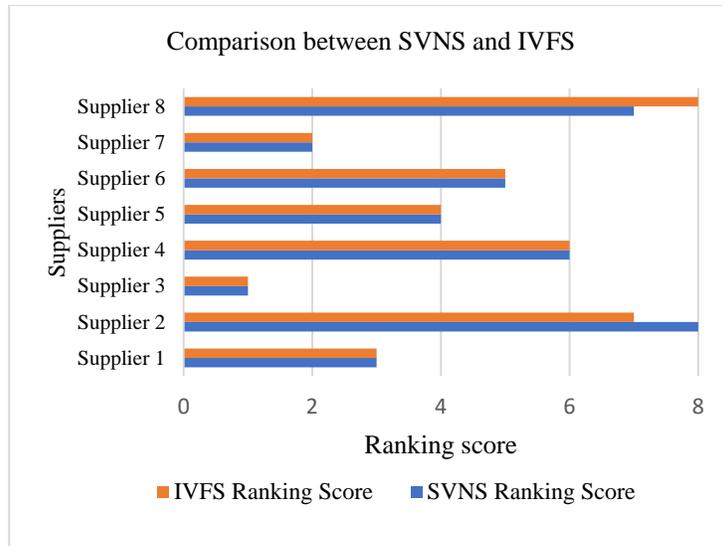

Figure 7. Comparison between SVNS and IVFS.

Table 11. Relationship between the Suppliers Resilience Performance and the Criteria.

| Supplier | Criteria | | | | | | | |
|---|---|---|---|---|---|---|---|---|
| | $C_1$ | $C_2$ | $C_3$ | $C_4$ | $C_5$ | $C_6$ | $C_7$ | $C_8$ |
| $S_1$ | P | N | N | N | P | N | N | P |
| $S_2$ | P | N | P | P | N | N | N | P |
| $S_3$ | P | P | N | P | N | N | P | P |
| $S_4$ | N | P | N | P | N | P | P | P |
| $S_5$ | P | P | P | P | N | N | P | N |
| $S_6$ | P | N | N | P | N | N | N | N |
| $S_7$ | P | N | P | P | P | N | P | P |
| $S_8$ | N | P | P | N | P | P | P | N |

Given the fact that performance of the candidate supplier(s) may differ on different criteria, the change in the importance of the criteria would change the overall performance of a candidate supplier concerning all the criteria. The relationship between the performance of the resilient suppliers and the criteria are summarized in Table 11 wherein P and N correspond to positive and Negative association. A positive association signifies that the resiliency property of the suppliers would benefit from the increase of the weight of that criterion, implying that the CC of the supplier would be higher if the importance of the criterion is increased. More conclusively, a positive association implies that a supplier possesses a good resilience performance on this criterion. A negative association between the supplier's performance and importance of the criteria has an opposite meaning.

We further compared the results obtained from the original membership based model and reliability-modified membership function based model and the original model and presented the outcome in Figure 8.

The result shows that the consideration of the reliability has slightly changed the CC of the suppliers, which indicates that the consideration of reliability influences the overall preferential ranking of suppliers.

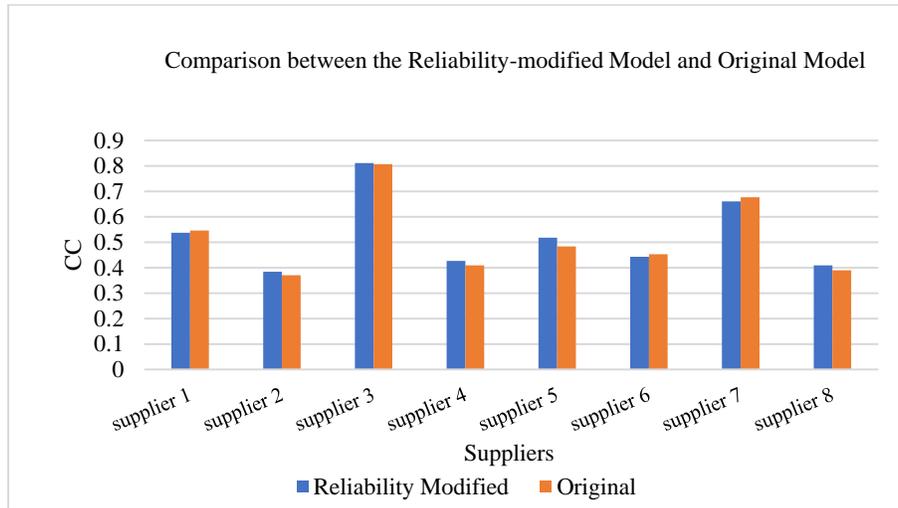

Figure 8. Comparison Between the Reliability-modified Model and Original Model

To further illustrate the importance of considering reliability-based membership function in MCDA framework, we adopt the method proposed in (Shahroudi & Tonekaboni, 2012) to perform another set of comparative analysis. Since, the existing method do not consider multiple decision makers, we use the mean values to integrate the differential evaluation given by multiple decision makers to make parallel comparison. To accommodate this comparison, crisp data 1 to 9 are assigned to linguistic terms to replace the SVNS. Based on the algorithm of the classical model, the generated results are presented in Table12.

Table 12. Results Generated from the classical model

| Supplier i | d+ | d- | cc | Ranking score |
|---|---|---|---|---|
| Supplier 1 | 0.02 | 0.03 | 0.62 | 4 |
| Supplier 2 | 0.01 | 0.03 | 0.68 | 1 |
| Supplier 3 | 0.02 | 0.03 | 0.60 | 5 |
| Supplier 4 | 0.02 | 0.03 | 0.64 | 2 |
| Supplier 5 | 0.02 | 0.02 | 0.47 | 6 |
| Supplier 6 | 0.03 | 0.02 | 0.46 | 7 |
| Supplier 7 | 0.01 | 0.03 | 0.63 | 3 |
| Supplier 8 | 0.03 | 0.02 | 0.41 | 8 |

The comparison of the result of our proposed MCDA model and the classical model are presented in Figure 9. From the figure, we observed significant changes in the ranking score of the candidate suppliers while adopting the reliability based SVNS approach compared to traditional MCDA approach. In classical approach, supplier 2 has the highest rank (represented by lowest ranking score), whereas the proposed model assigns supplier 2 the lowest rank (represented by highest ranking score). Similar result was obtained for supplier 4, the proposed model favors this supplier a lot less compared to the classical model. For rest of the suppliers, slightly different outcomes were observed; higher ranking for suppliers in proposed model were obtained when compared with the ranking given by the classical model. Therefore, such comparison with classical model clearly demonstrates that consideration of reliability modified SVNS with TOPSIS better handle the uncertainty inherent in the imprecise and incomplete preference given by the multiple decision makers.

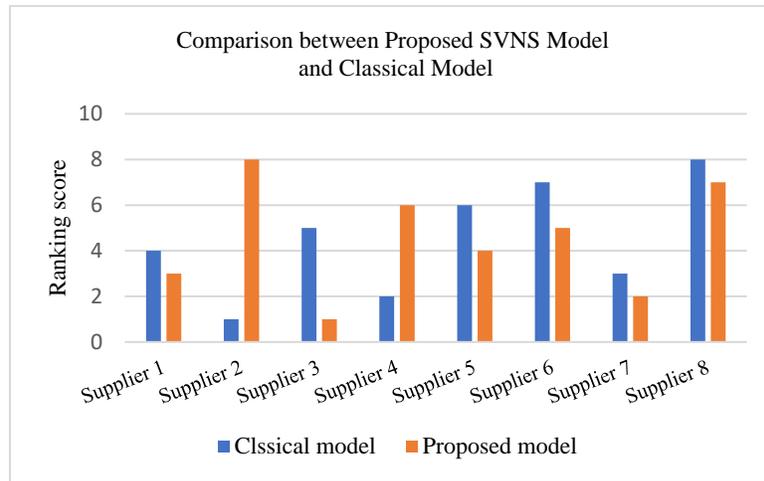

Figure 9. Comparison between Proposed SVNS Model and Classical Model.

**4.2 Sensitivity Analysis**

In order to generate managerial insights on the relative importance of different criteria in selecting a resilient supplier selection, we perform sensitivity analyses to see how the decision changes as the weights of various criteria do change. In real world, the requirements of the decision-makers who need to select the optimal supplier differ significantly due to their diverse preferences on several criteria. For example, some of the decision-makers may care more about buffer capacity while the others' satisfaction would be fulfilled only when the alternative shows a superior performance on responsiveness. That is, the ranking of the supplier(s) may change while different weights are assigned to different criteria. To test whether our proposed algorithm is capable of explaining this, we conduct a sensitivity analysis with respect to the variation in the weight of criteria to observe the corresponding change in the CC and the final list of preference.

The weight in the form of SVNS is summarized in Table 4. For SVNS, the higher the importance of a criterion, the larger the truth value. So, we can adopt the truth value to represent the SVNS weight to do the sensitivity analysis. Considering C7 as example, the SVNS for $C_7$ is (0.34, 0.76, 0.79) and assuming the representative weight for $C_j$ is $\alpha_j$, then we have $\alpha_7 = 0.34$. To check the impact of the value of $\alpha_7$ on the final ranking score (CC), we would slightly change the value of $\alpha_7$. Because the weights are calculated from the original linguistic data based on Table 13, we can gradually change the original linguistic data on $C_7$ to obtain the $\alpha_7$ we desired. The original linguistic data for $C_7$ and corresponding value of $\alpha_7$ are summarized in Table 13.

Table 13. Original Linguistic Data for Criteria 7 ($C_7$).

| Data Set | Original Linguistic Data for $C_7$ | | | | | | | | | | $\alpha_7$ |
| --- | --- | --- | --- | --- | --- | --- | --- | --- | --- | --- | --- |
| DMs | $DM_1$ | $DM_2$ | $DM_3$ | $DM_4$ | $DM_5$ | $DM_6$ | $DM_7$ | $DM_8$ | $DM_9$ | $DM_{10}$ | |
| Original Data Set | UI | UI | VUI | UI | VUI | VUI | M | M | UI | UI | 0.34 |
| Data Set 1 | VUI | VUI | VUI | VUI | VUI | VUI | VUI | VUI | VUI | VUI | 0.1 |
| Data Set 2 | VUI | VUI | VUI | VUI | VUI | VUI | UI | UI | UI | UI | 0.2 |
| Data Set 3 | VUI | VUI | VUI | UI | UI | VUI | M | M | UI | UI | 0.3 |
| Data Set 4 | M | M | UI | UI | VUI | VUI | M | M | UI | UI | 0.4 |
| Data Set 5 | M | M | UI | M | UI | UI | I | I | UI | UI | 0.5 |
| Data Set 6 | I | I | UI | UI | UI | UI | I | I | UI | M | 0.6 |
| Data Set 7 | I | I | M | I | M | M | VI | I | M | M | 0.7 |
| Data Set 8 | VI | VI | I | I | UI | M | VI | VI | I | I | 0.8 |
| Data Set 9 | VI | VI | VI | VI | VI | VI | VI | VI | VI | VI | 0.9 |

While we change the value of $\alpha_7$, the SVNS weight of other criteria would remain the same based on (2.13). However, as the original performance measurement will be multiplied by the weight to get the weighted performance measurement, the suppliers who have better performance on $C_7$ would benefit from the increased weight, leading to a higher CC which, eventually will change the preferential order of the alternative suppliers.

Fig 10 shows the sensitivity analysis on $C_7$. In the left picture, we show all the supplier, while in the right we show only the suppliers with the highest-ranking score for these criteria, e.g., supplier 3 and supplier 7.

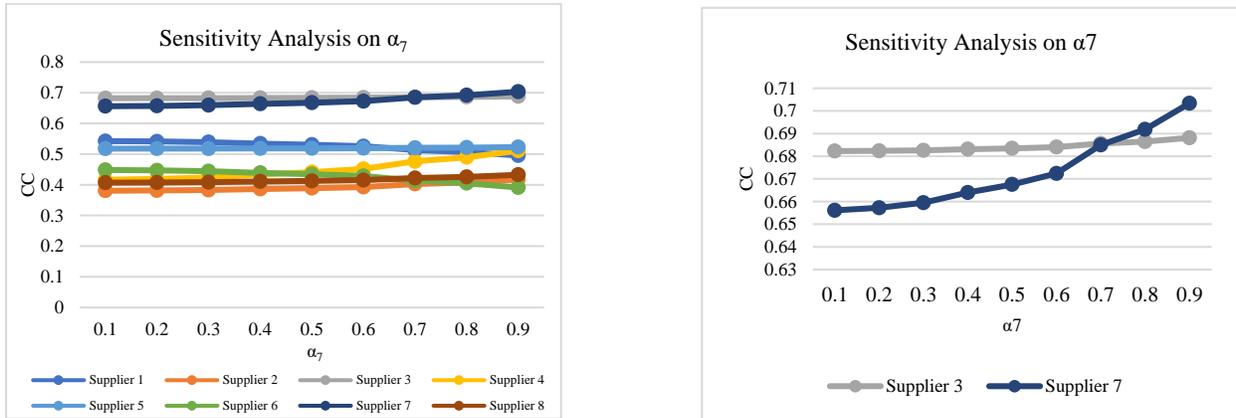

Figure 10. Sensitivity Analysis on Criteria 7 ($C_7$) for the Proposed Model.

We found that, when $\alpha_7 < 0.7$, the optimal supplier is supplier 3; when $\alpha_7 \geq 0.7$, supplier 3 loses its priority over supplier 7. We must mention that, as the $\alpha_i$ are independent from each other, we only focus on one $\alpha_i$ at one time. Particularly, in doing the sensitivity analysis against one criterion, only the weight of that criterion is adjusted while keeping the weight of other criteria constant. We can do the sensitivity analysis on every $\alpha_i$ separately to check their influence on the final rank. The results are summarized in Table 14.

Table 14. Result of the Sensitivity Analysis.

| $\alpha_j$ | Range | Optimal Supplier | Range | Optimal Supplier |
|---|---|---|---|---|
| $\alpha_1$ | (0.1,0.9) | $S_3$ | - | - |
| $\alpha_2$ | (0.1,0.9) | $S_3$ | - | - |
| $\alpha_3$ | (0.1,0.72) | $S_3$ | (0.72,0.9) | $S_7$ |
| $\alpha_4$ | (0.1,0.8) | $S_3$ | (0.8,0.9) | $S_7$ |
| $\alpha_5$ | (0.1,0.9) | $S_3$ | - | - |
| $\alpha_6$ | (0.1,0.75) | $S_3$ | (0.75,0.9) | $S_7$ |
| $\alpha_7$ | (0.1,0.7) | $S_3$ | (0.7,0.9) | $S_7$ |
| $\alpha_8$ | (0.1,0.75) | $S_3$ | (0.75,0.9) | $S_7$ |

To show the superiority of our proposed model, we conducted a sensitivity analysis on the criteria weight for the classical model (Shahroudi & Tonekaboni, 2012) as well. Since the criteria weights in the classical model are generated from the suppliers' performance measurement matrix which are not given, we slightly modified the classical approach by assigning an additional coefficient $\alpha$ to adjust the importance of a criteria given by a decision maker. The result of the sensitivity analysis is summarized in the Figure 11, which shows that regardless of the change in the criteria weight, supplier 2 always remains optimal. However, from a realistic point of view, it is expected that the optimal supplier ranking should change with the change in criteria importance, requiring the decision-making framework to be sensitive to the change in importance of the evaluation criteria. Depicted in figure 10, the proposed algorithm can generate alternate optimal supplier based on the different importance of criteria, however, the traditional approach fails to respond to that change in criteria weight. Therefore, sensitivity analysis demonstrated the superiority of the proposed algorithm over traditional approach in terms of its strength to capture the decision maker's preference regarding an evaluation criteria.

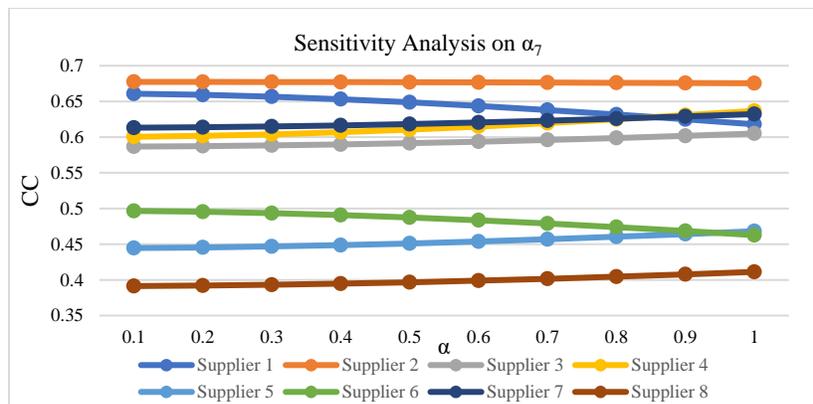

Figure 11. Sensitivity Analysis on Criteria 7 ($C_7$) for the Classical Model.

From managerial perspective, determining the impact of different criteria as well as the quantification of the influence of relative importance of the criteria on decision is of utmost importance to the decision makers. Based on the changes in the internal business strategies, decision makers most often set different priority for different criteria that are considered to assess the resiliency of suppliers. A comprehensive evaluation of available suppliers considering such priority changes is crucial for planning better strategic sourcing decisions from resilience perspective. This, in turn helps management take informed decisions, which can potentially mitigate the risk of disruption both in supply- and demand-side, leading to enhanced resilience of the entire supply chain. Therefore, through this sensitivity analyses, we showed how our proposed decision-making framework can quantify the impact of differential priority given by multiple decision makers on different evaluation criteria, and carries the potential to provide decision makers with valuable managerial insights that are instrumental to ensure resilience against disruptions.

## 5. Conclusion

Resilient suppliers are critical for strategic manufacturers to help mitigate the negative impact of any natural or man-made disruption, leading to enhanced capability of supply chain to sustain both pre and post disaster impact. A great deal of uncertainty both in the supply- and demand-side of a supply chain is possible to manage in an effective way via selecting a resilient supplier. Thus, the focus of this study lies in the development of a decision-making framework to facilitate the evaluation of a set of available suppliers from resilience perspective. Such comprehensive evaluation process entails the assessment of uncertainty inherent in imprecise DRI given by a group of decision makers on both the numerical and linguistic criteria. This is an issue, which has received limited attention in traditional MCDA based supplier selection literatures. To that end, we adapt and extend SVNS and IVFS techniques to consider quantitative and linguistic evaluation criteria simultaneously in the proposed decision-making algorithm. To consider unreliability of the information and thus to better capture the uncertainty, we calculate reliability indices to obtain reliability adjusted membership functions. With the help of reliability-based membership degree, the crisp data on numerical criteria are transferred to aggregated SVNS or IVFS decision matrix, which are then combined with the decision matrix on linguistic criteria and each supplier is ranked by the TOPSIS tool. The comparative results generated from these two approaches (SVNS and IVFS) verifies the effectiveness of the proposed algorithm. We also perform a sensitivity analysis to demonstrate how the priority of an evaluation criterion impacts the preferential ranking of the suppliers. Results showed that the proposed decision-making framework can suggest alternate optimal supplier when the importance of any criteria changes. Being sensitive to such changes in criteria importance, the proposed framework can empower the decision makers to evaluate different suppliers and better plan the sourcing decisions to enhance resilience. Thus, the proposed decision-making framework is anticipated to be used as an effective and reliable tool for supply chain stakeholders to evaluate multiple resilient suppliers considering several conflicting criteria with imprecise decision relevant information. One of the few limitations of this study is that out of many other fuzzy techniques, we only considered SVNS and IVFS; thus, providing opportunities for future studies that can potentially investigate the feasibility and effectiveness of Triangular Interval Type-2 Intuitionistic Fuzzy Set (TIT2IFS) with weighted averaging (Garg, 2016; Singh & Garg, 2018), Dual Hesitant Fuzzy Soft Set (DHFSS) augmented with weighted correlation coefficient (Arora & Garg, 2018) measure as these techniques showed promising performance in handling uncertainties and vagueness while evaluating alternatives. In addition, further research may also extend this current work to develop a method that can optimize the weight of decision maker while aggregating the numerical data set.

**List of Figures**

Figure 1. Illustration for the Triangular Interval Valued Fuzzy Number.

Figure 2. Frame of Discernment.

Figure 3. Illustration of Static Reliability Index.

Figure 4. Illustration of Dynamic Reliability Index.

Figure 5. Illustration of Fuzzification.

Figure 6. Framework of the Proposed MCDA algorithm.

Figure 7. Comparison between SVNS and IVFS.

Figure 8. Comparison Between the Reliability-modified Model and Original Model.

Figure 9. Comparison between Proposed SVNS Model and Classical Model.

Figure 10. Sensitivity Analysis on Criteria 7 ($C_7$).

Figure 11. Sensitivity Analysis on Criteria 7 ($C_7$) for the Classical Model.